\documentclass[journal]{IEEEtran}
\usepackage{graphicx}
\usepackage{subfigure}
\usepackage{mathrsfs}
\usepackage{amssymb}
\usepackage{booktabs}
\usepackage{bbm}
\usepackage{array}
\usepackage{color}
\usepackage{hyperref}
\hypersetup{
	breaklinks=true,
	letterpaper=blue,
	bookmarks=false     
}
\ifCLASSINFOpdf
\else
\fi
\begin{document}
	%
	\title{Audio-visual Representation Learning for \\Anomaly Events Detection in Crowds}
	%
	%
	%
	
	\author{
		Junyu~Gao,~\IEEEmembership{Member,~IEEE,}
		~Maoguo~Gong,~\IEEEmembership{~Senior Member,~IEEE,}
		and Xuelong Li,~\IEEEmembership{~Fellow,~IEEE}
		\thanks{

			J. Gao is with the Academy of Advanced Interdisciplinary Research,
			Xidian University, Xi’an 710071, Shaanxi, China, and the School of Artificial Intelligence, Optics and Electronics (iOPEN), Northwestern Polytechnical University, Xi'an {\rm 710072}, P. R. China. E-mail: gjy3035@gmail.com.
			
			M. Gong is with the Key Laboratory of Intelligent Perception and Image Understanding of Ministry of Education, International Research Center for Intelligent Perception and Computation, Xidian University, Xi’an 710071, Shaanxi, China. E-mail: gong@ieee.org.
			
			X. Li is with the School of Artificial Intelligence, Optics and Electronics (iOPEN), Northwestern Polytechnical University, Xi'an {\rm 710072}, P. R. China. E-mail: li@nwpu.edu.cn.
			
			Corresponding author: X. Li.
		}
	}
	
	%
	%

	\markboth{Journal of \LaTeX\ Class Files,~Vol.~14, No.~8, August~2015}%
	{Shell \MakeLowercase{\textit{et al.}}: Bare Demo of IEEEtran.cls for IEEE Journals}
	%



	\maketitle
	
	\begin{abstract}
		In recent years, anomaly events detection in crowd scenes attracts many researchers' attentions, because of its importance to public safety. Existing methods usually exploit visual information to analyze whether any abnormal events have occurred due to only visual sensors are generally equipped in public places. However, when an abnormal event in crowds occurs, sound information may be discriminative to assist the crowd analysis system to determine whether there is an abnormality. Compare with vision information that is easily occluded, audio signals has a certain degree of penetration. Thus, this paper attempt to exploit multi-modal learning for modeling the audio and visual signals simultaneously. To be specific, we design a two-branch network to model different types of information. The first is a typical 3D CNN model to extract temporal appearance feature from video clips. The second is an audio CNN for encoding Log Mel-Spectrogram of audio signals. Finally, by fusing the above features, the more accurate prediction will be produced. We conduct the experiments on SHADE dataset, a synthetic audio-visual dataset in surveillance scenes, and find introducing audio signals effectively improves the performance of anomaly events detection and outperforms other state-of-the-art methods. Furthermore, we will release the code and the pre-trained models as soon as possible. 
	\end{abstract}
	

	%
	\IEEEpeerreviewmaketitle

	\section{Introduction}
	\label{intro}
	%
	%
	%
	%
	\IEEEPARstart{C}{rowd} analysis is an essential task in the field of public safety including crowd counting \cite{gao2019learning,zhou2020adversarial}, localization \cite{abousamra2020localization,9347744,gao2020learning}, anomaly events detection \cite{thida2013laplacian,lin2021learning}, flow/motion analysis \cite{ali2007lagrangian,rao2015crowd}, segmentation \cite{kok2016grcs,wang2021pixel}, group detection \cite{mazzon2013detection,li2017multiview},  \emph{etc.} Especially, anomaly events detection is a fundamental task for safety warning in crowd scenes. Timely alarming of anomaly events that are occurring is essential to ensure public safety. By our observation, we find that when an abnormal event occurs, it is often accompanied by some special sounds. Unfortunately, since the common surveillance cameras are not equipped with audio collection recorders, many researchers only pay attention to the visual information and ignore the audio signal. However, there are many cases that occur in the blind spot of surveillance cameras, resulting in the security system not being able to recognize and warn for police or others in advance. Thus, in this paper, we attempt to explore the effect of ambient sound on the task of anomaly event detection.
	
	\begin{figure*}[h]
		\centering
		\includegraphics[width=\linewidth]{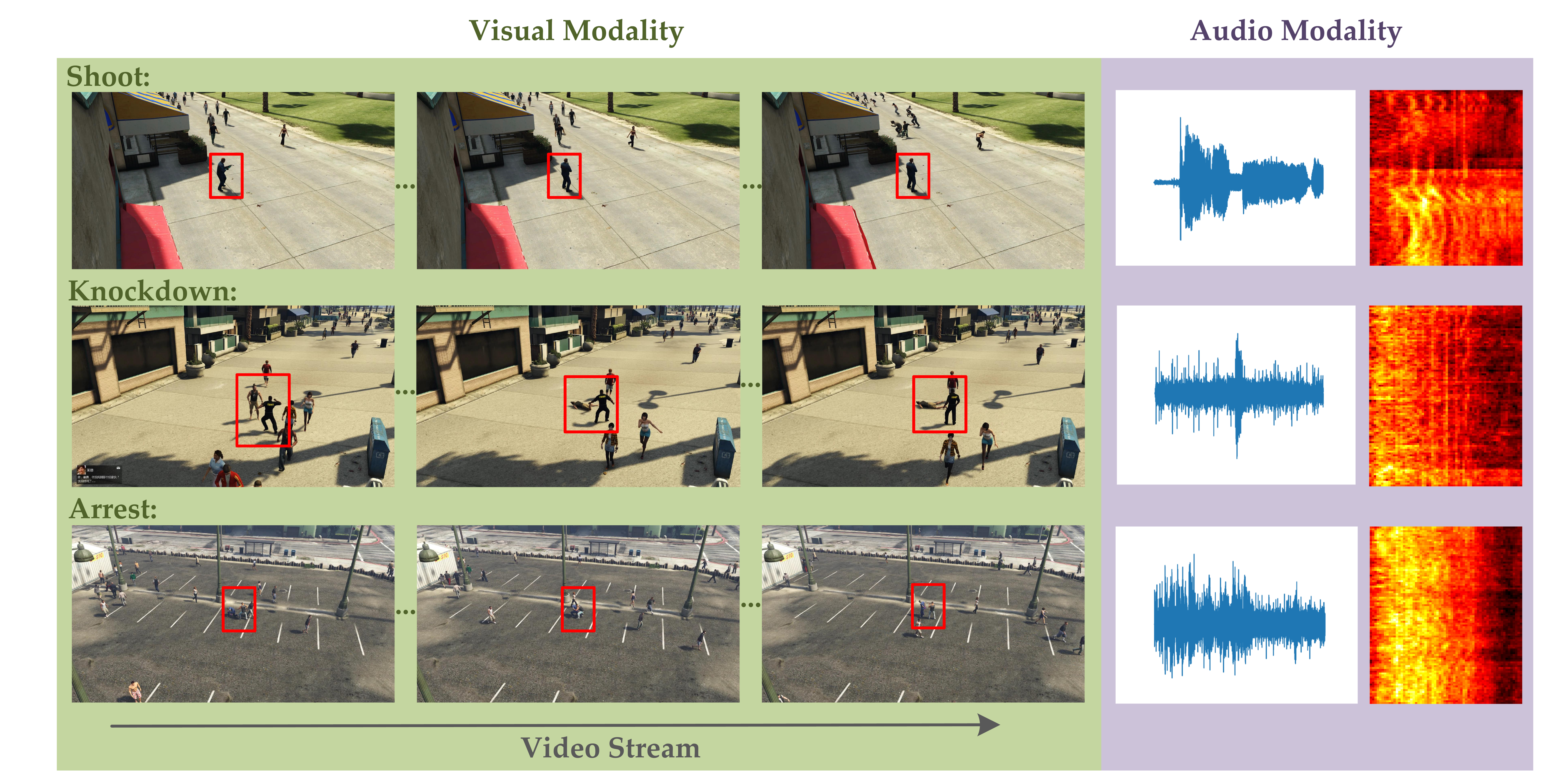}
		\caption{Some typical samples in SHADE dataset.  The \textcolor{green}{green box} contains three groups of image sequences of Shoot, Knockdown, and Arrest, of which \textcolor{red}{red boxes} in frames highlight the incident location. The \textcolor{magenta}{magenta box} visualizes the audio signals and Mel-Spectrograms for the above crowd scenes.}
		\label{fig1}
	\end{figure*}
	
	With the development of deep learning on computer vision, there are many phenomenal works in the fields of image recognition \cite{krizhevsky2012imagenet,simonyan2014very,he2016deep}, temporal modeling \cite{karpathy2014large,DBLP:journals/pami/BinzhaoRSGN,liVideoDistillation2021}, and audio signals processing \cite{hershey2017cnn,gemmeke2017audio,hu2020curriculum}. In the field of crowd analysis, most of the existing work focuses on the research of still images \cite{wan2019residual,gao2020pcc} and image sequences \cite{sultani2018real,Singh_2018_CVPR_Workshops}, namely visual crowd analysis. However, currently there is only one method that combines visual and auditory information \cite{hu2020ambient}, which proposes a multi-modal learning to encode still images and ambient sounds simultaneously. The method significantly reduces the estimation errors for crowd counting in extreme conditions. Inspired this, we investigate common abnormal events under video surveillance in crowd scenes, such as fighting, fleeing, chasing, chaos, \emph{etc.} and find that the crowd can make some special sounds, which may aid model extract more discriminative features to recognize the event. Fig. \ref{fig1} illustrates the videos and the corresponding audio data of some typical samples in SHADE dataset \cite{lin2021learning}. From it, there are many differences among different events' ambient sound. Besides, the surveillance camera may be blocked by obstacles and cannot capture effective content. In this case, the sound signal shows its unique advantage: it can bypass obstacles and be collected by recording equipment.
	
	Based on the aforementioned observation, this paper propose a multi-modal learning to extract features from image sequences and audio data for anomaly event detection in crowd scenes. It named as ``Audio-visual Representation Learning'' (AVRL for short), which consists of two streams: a Residual 3D CNN \cite{DBLP:conf/iccvw/HaraKS17} is developed on image clips and a audio CNN on processed audio information represented by Log Mel-Spectrogram algorithm \cite{DBLP:conf/icassp/HersheyCEGJMPPS17}. After obtaining two types of features, a feature-level fusion mechanism is presented to integrate them and the event class is directly predicted.
	
	In summary, this paper has three-fold contributions:
	
	\begin{enumerate}
		\item[1)] Propose a new framework, Audio-visual Representation Learning (AVRL), to extract appearance and sound features from image sequences and the corresponding audio signals. 
		
		\item[2)] Present an effective scheme to fuse the spatial-temporal 3D-CNN's features and temporal audio features, which can achieve the balanced trade-off between visual and audio representations and reaches better anomaly detection accuracy. 
		
		\item[3)] Outperform the state-of-the-art method on the anomaly events detection task in crowd scenes.
		
	\end{enumerate}
	
	The rest of this paper is organized as follows: Section \ref{related} briefly lists and reviews the related works about crowd analysis and multi-modal learning; Section \ref{method} presents the proposed AVRL method for anomaly event detection; Section \ref{exp} and \ref{dis} conduct the extensive experiments and further analyze the the key settings of the method. Finally, this work is summarized in Section \ref{concl}. 
	
	\begin{figure*}[h]
		\centering
		\includegraphics[width=\linewidth]{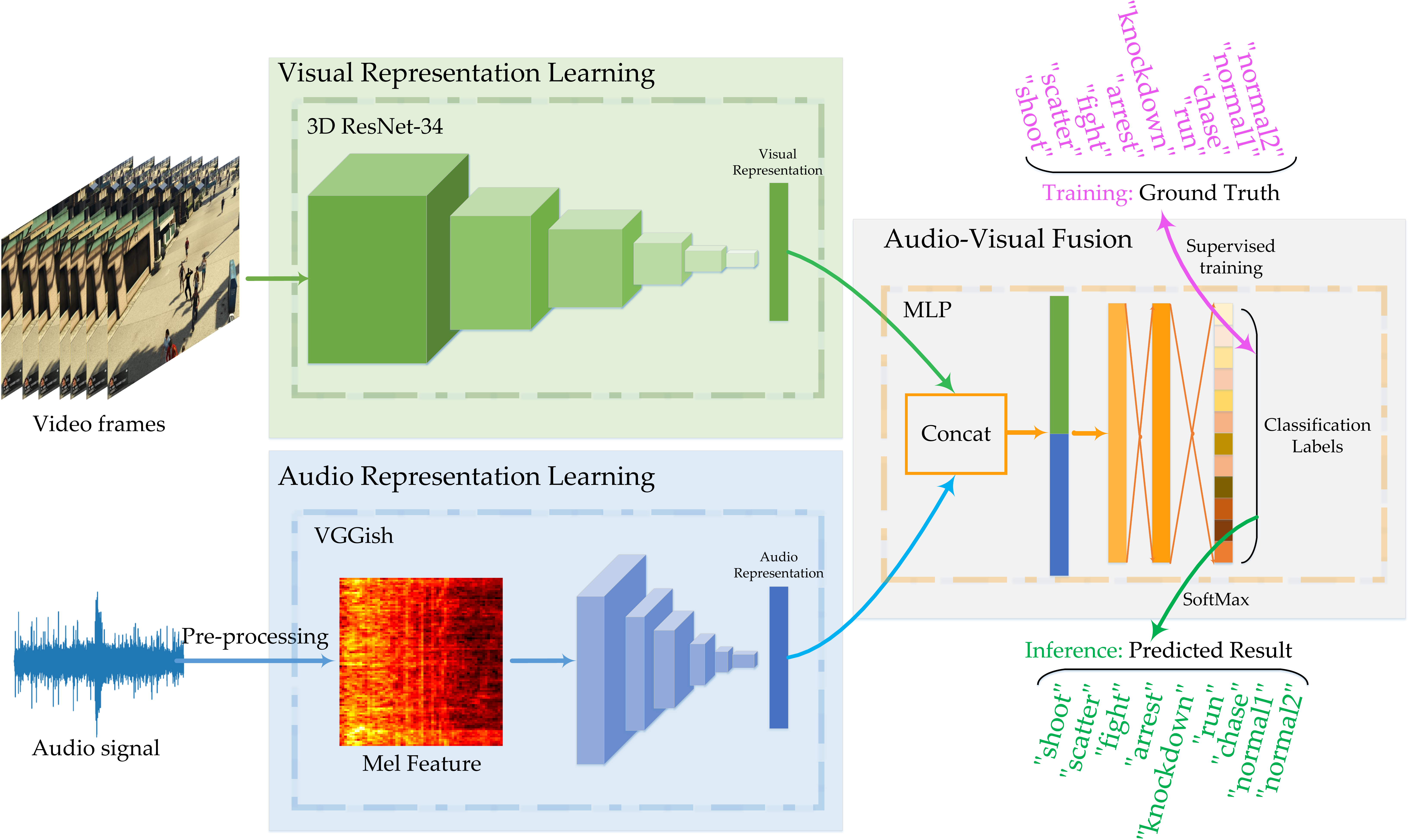}
		\caption{\textbf{AVRL} for abnormal events detection in crowd scenes. The details of 3D ResNet and VGGish is the same as \cite{DBLP:conf/iccvw/HaraKS17} and \cite{DBLP:conf/icassp/HersheyCEGJMPPS17} respectively.}
		\label{AVRL}
	\end{figure*}
	
	\section{Related Work}
	\label{related}
	This section briefly reviews the related works about research fields in this paper: abnormal events detection in crowd scenes and audio-visual representation learning. 
	
	\subsection{Abnormal Events Detection in Crowds}
	The recent anomaly detection methods can be divided into two main categories: local analysis and global analysis. As for local analysis methods, the locations where abnormal events occur are detected. The typical local analysis methods include \cite{DBLP:conf/mm/ChengCF13, DBLP:journals/tip/DuZZT16, DBLP:journals/informaticaLT/ChenNYA18, DBLP:journals/tgrs/DuRWZ19}, \emph{etc.} These methods detect the abnormal objects, such as trucks or running people in crowds and localize the position where anomalies occur. Liao \emph{et al.} \cite{DBLP:conf/icig/LiaoXSFD11} use video descriptors to detect the fighting event in video frames. The other category, global analysis, usually analyze the whole video clips and predict whether there are abnormal events in the video clips. The global analysis methods have three typical schemes: trajectory points based, optical flow based, and classification based. \cite{DBLP:conf/cvpr/MehranOS09, DBLP:journals/vlc/CuiLX14, DBLP:journals/tcsv/ZhangQJYH15} utilize the object trajectories extracted from crowd video clips for anomaly detection. Helbing \emph{et al.} \cite{PhysRevE.51.4282} proposed is a typical  trajectory points based method named social force model for crowd interaction description. Inspired by \cite{PhysRevE.51.4282}, \cite{DBLP:conf/cvpr/MehranOS09} and \cite{DBLP:journals/tcsv/ZhangQJYH15} introduce the social force model into the social events analysis in videos. Cui \emph{et al.} \cite{DBLP:journals/vlc/CuiLX14} attempt to cluster trajectories by Fuzzy C-Means Clustering \cite{DBLP:journals/pami/CannonDB86} and predict the category of input trajectory through clustering. Du \emph{et al.} \cite{DBLP:journals/tgrs/DuRWZ19} propose a change detector named DSFA, and it utilizes two symmetric streams and slow feature analysis module to acquire better performance on changes in remote sensing imagery. 
	
	Because the definition of abnormal events in statistic based methods is ambiguous and inchoate, the VSD \cite{DBLP:journals/mta/DemartyPSG15} and UCF-Crime \cite{DBLP:conf/cvpr/SultaniCS18} are proposed for abnormal events detection in crowd scenes, containing 7 and 13 different categories about anomalies respectively. The video clips in these two datasets are gained from movies and videos on internet. \cite{DBLP:journals/mta/MironicaDIS16} is a typical method for video classification with VSD benchmark. In the meantime, \cite{DBLP:journals/corr/RabieeHMNMS16} estimates the crowd emotion for abnormal behavior analysis and understanding on dataset in \cite{DBLP:conf/avss/RabieeHMKNM16}. The datasets mentioned above define the abnormal events in crowd detailedly, leading the abnormal detection task to the video classification task. Therefore, we deploy the recent video classification methods into the anomaly detection and receive excellent results. 
	
	\subsection{Audio-Visual Multi-Modal Learning}
	The joint audio-visual multi-modal learning attempts to the learn the representation from visual and auditory modalities for special tasks. The early researches on this field concentrate on the speech recognition task \cite{DBLP:journals/tmm/DupontL00}. The motivation of these works is that the visual modality provides auxiliary information for audio analysis with noisy. A typical research on audio-visual multi-modal learning is to model the movement of face and mouth through the videos and corresponding audio signals \cite{DBLP:conf/icml/NgiamKKNLN11, DBLP:conf/cvpr/HuLL16}. Similarly, the multi-modal learning is introduced into other tasks \cite{DBLP:journals/tmm/KettebekovYS05, DBLP:journals/tmm/ZengTLHPRL07}. 
	
	In recent years, the  audio-visual multi-modal learning is utilized in more general scenes. Owens \emph{et al.} \cite{DBLP:conf/eccv/Owens0MFT16, DBLP:journals/ijcv/OwensWMFT18} try to transfer the knowledge from audio learning to visual learning. Arandjelovic \emph{et al.} \cite{DBLP:conf/iccv/ArandjelovicZ17} analyze the video through audio-visual relations. Meanwhile, this work is employed to sound localization \cite{DBLP:conf/cvpr/HuNL19, DBLP:journals/corr/abs-2001-09414} and audio-visual separation \cite{DBLP:conf/iccv/GanZCC019}. 
	
	This paper is inspired by these audio-visual multi-modal learning methods and we attempt to  introduce the audio modality into the traditional visual-based abnormal detection methods. The anomaly detection in crowd scenes benefits from special ambient sounds in some events.

	\section{Our Approach}
	\label{method}
	The proposed framework AVRL contains three main modules: visual representation learning, audio representation learning, and audio-visual representation fusion, showing in Fig. \ref{AVRL}. Compared with the conventional video anomaly detection method, our framework has the unique audio representation learning, and audio-visual representation fusion modules for abnormal events detection in crowd analysis. As mentioned above, the motivation of our framework is that some abnormal events have the extremely special ambient sounds, such as shooting, scattering, \emph{etc.} and these discriminating ambient sounds is beneficial for abnormal events recognition. For example, when the shooting rampage occurred, some people near the crime scene may not see the shooter at all, but the loud gunfire helps the people in judging the abnormal event. To introduce this judgment capacity, we design a audio representation learning module in the traditional video abnormal events detection framework, gaining the \emph{Audio-visual Representation Learning  (AVRL)} framework.
	
	\subsection{Visual Representation Learning}
	Compared with the images, videos have inherent temporal correlation, and video data have extra time dimension. In order to extract the temporal information, 3D convolutional filter, a 2D convolutional filter with time dimension, is proposed and introduced into the vision task on videos. Thus, the video representation learning module in AVRL framework uses 3D-ResNet \cite{DBLP:conf/iccvw/HaraKS17} to extract the visual representation with temporal correlation among video frames. 
	
	Here, the input videos with $l$ frames are sampled to a fixed length $T$ ($T=50$ in this paper) with a fixed sample interval $\lfloor l/T \rfloor$. This fixed length frame sequence $\mathcal{V} \in \mathbb{R}^{C\times T \times H\times W}$ is passed to the visual representation learning module $\mathcal{F}_{vid}$. The visual representation $f_{v}$ is extracted with the equation:
	\begin{equation}
	f_{v}=\mathcal{F}_{vid}(\mathcal{V}),
	\end{equation}
	where $v_{feat} \in \mathbb{R}^{C\times 1\times 1}$ is the finally output of 3D-ResNet-34, and $C$ is the number of channels ($C_v=512$ in 3D-ResNet).
	
	In vanilla 3D-ResNet, the representation $v_{feat}$ is mapped to logits for events classification, achieving the abnormal events detection. In our proposed AVRL framework, this video representation do not use for classification directly, it will be merged with audio representation to realize the multi-modal learning system. 
	
	\subsection{Audio Representation Learning}
	With the development of deep learning, there are some CNN-based methods \cite{DBLP:conf/icassp/HersheyCEGJMPPS17} have applied on audio analysis successful. The CNN-based method are usually utilized to the image data due to the characteristic of convolutional layer. However, the audio signals from videos only has one-dimension, and a audio signal preprocess is necessary for the CNN-based audio representation learning. 
	
	In this paper, Log Mel Spectrogram (LMS) for audio preprocess is used in audio representation learning module, because the LMS features from audio is used in CNN-based audio task generally.  According to the VGGish \cite{DBLP:conf/icassp/HersheyCEGJMPPS17}, the input audio signal from video is first resampled with 16kHz, and then the time-frequency map of resampled audio signal is gain through a Short Time Fourier Transform (STFT) with Hann window, and the details of preprocess are same as VGGish. The LMS feature $\mathcal{A} \in \mathbb{R}^{H\times W}$ has two-dimensiom, and the CNN-based methods can easily extract the hidden feature from audio signal. In AVRL, we use VGGish as audio representation learning module $\mathcal{F}_{aud}$. The audio representation extraction is defined as:
	\begin{equation}
	f_{a}=\mathcal{F}_{aud}(\mathcal{A}),
	\end{equation}
	where $f_{a} \in \mathbb{R}^{C_a\times 1}$ is the audio representation, and $C_a=512$ in VGGish.
	
	\subsection{Audio-visual Representation Fusion}
	The traditional abnormal events detection methods only analyze the feature from videos, but some anomaly is not noticeable in crowd scene. Supposing that two people are fighting in a congested crowd, this fight event is not obvious in vision due to the dense crowd, but if the people in fight are roaring or crying, the fight event will be judged easily. Under this circumstance, the fusion of audio and visual representation has the advantages for events detection in crowd analysis. To detect abnormal events effectively, we fuse the audio and visual representation by audio-visual representation fusion module. In order to select the better fusion methods, we design several fusion strategy and compared their performances in Section \ref{comp_fusion}. Finally, we select a simple and effective fusion method for AVRL framework.
	
	Based on the visual/audio representation learning above, we fuse the representations through feature concatenation, and then pass this multi-modal feature into classification network $\mathcal{F}_{cls}$ to detect whether abnormal events occur in the video. The formula is shown here: 
	\begin{equation}
	p=\mathcal{F}_{cls}(Cat(f_{v},f_{a})),
	\end{equation}
	where $p \in \mathbb{R}^{N\times 1\times 1}$ is the event category prediction, $N$ is the number of categories in SHADE, $Cat$ denotes the feature concatenate manipulation. This simple fusion module achieve the better performance in comparison with other fusion strategy we designed, and the details of different fusion strategy is shown in Section \ref{comp_fusion}. The classification network $\mathcal{F}_{cls}$ in audio-visual representation fusion module is a simple fully connected network. 
	
	Intuitively, the audio-visual representation fusion module combines the visual representation and audio representation, and extracts the comprehensive feature with fully connected network for abnormal events classification. With the auxiliary information from audio, the audio-visual combined representation is more discriminating compared with traditional visual methods. Although the fusion module of AVRL framework is quite simple, the experiment results illustrate that this simple strategy is effective in abnormal events detection, and even surpasses more complex fusion methods. 
	
	\subsection{Loss Function}
	Anomaly events detection in crowd scenes is a video classification task. In order to find the anomaly events, the model should judge whether the action in video is belong to abnormal events, such as shoot, scatter, fight, \emph{etc.} Given a video with category $y$ and a predict category $p$, we selected Cross Entropy Loss, a typical loss function for classification task. The loss function $\mathcal{L}$ is defined as:
	\begin{equation}
	\mathcal{L}= -\frac{1}{M} \sum_{i=1}^{M} \sum_{j=1}^{N} y_{ij}log(p_{ij}),
	\end{equation}
	where $M$ is the mini-batch size, $N$ denotes the number of categories.
	
	\subsection{Other Details}
	We select the 3D-ResNet-34 pre-trained on UCF-101 \cite{UCF} as the backbone for visual representation learning on video data. 3D-ResNet is a widely used deep learning model for video vision task. 3D-ResNet replace the traditional 2D convolutional layer in ResNet with 3D convolutional layer, introducing the temporal correlation in CNN. 3D convolutional filter adds an extra length dimension in 2D convolutional filter. 
	
	To be specific, suppose $F^{i,j}_{x,y,z}$ is the value at position $(x,y,z)$ on $j^{th}$ feature map from the output of $i^{th}$ 3D convolutional layer, and the input feature of $i^{th}$ layer has $C$ channels. The 3D convolutional filter has $H$ height, $W$ width, and $L$ length. The 3D convolution is shown as follow:
	\begin{equation}
	F^{ij}_{x,y,z}= b^{ij}+\sum^{C}_{c=1} \sum^{H}_{h=1} \sum^{W}_{w=1} \sum^{L}_{l=1} w^{ij}_{h,w,l} \cdot F^{(i-1)m}_{x+w,y+h,z+l} \ ,
	\end{equation}
	where $w^{ij}$, and $b^{ij}$ are the weight and bias of convolutional filter respectively.
	
	In addition, we select a common audio analysis CNN, VGGish pre-trained on AudioSet \cite{AudioSet}, to learn the audio representation. The PCA postprocess of VGGish is removed from our audio representation learning module. Due to the discrepancy of audio length in SHADE dataset, the LMS features have different shapes. We impose Global Average Pooling (GAP) along channel on the output of audio representation, so that the audio representation $f_{a}$ has a unified size $C_a\times 1$.
	
	The fused audio-visual representation is passed to a fully connected network for event categories classification. We design a simple fully connected network $\mathcal{F}_{cls}$, and the architecture of  $\mathcal{F}_{cls}$ is illustrated in Table \ref{fc}:
	
	\begin{table}[h]
		\centering
		\small
		\caption{Classification network in Audio-visual representation fusion module.}
		\label{fc}
		\begin{tabular}{lccc}
			\toprule[1pt]
			&\textbf{Layer}& \textbf{Input Size} & \textbf{Output Size} \\
			\hline
			0&Linear&1024&512 \\
			1&ReLU&512&512 \\
			2&Linear&512&256 \\
			3&ReLU&256&256 \\
			4&Linear&256& \# of categories \\
			5&Classification \\
			\toprule[1pt]
		\end{tabular}
	\end{table}
	
	\section{Experiments}
	\label{exp}
	
	The experiments are conducted on SHADE dataset. The experimental details and discussions are shown in this section.
	
	\subsection{Evaluation Metrics}
	Abnormal events detection can be considered as a video classification task, so that the classification accuracy is the core metric for this task. The Top-1 Accuracy is gauged to compare the performances among different methods:
	\begin{equation}\label{indicator}
	Acc_i = \frac{1}{N} \sum_{j=1}^{M} \mathbbm{1}\{y_i,p_i\},
	\end{equation}
	\begin{equation}{}
	\mathbbm{1}\{y_i,p_i\}=\left\{ \begin{array}{rcl}
	1 & \mbox{if} &  p_i=y_i \\
	0 & \mbox{if} & p_i\neq y_i
	\end{array} \right. ,
	\end{equation}
	where i indicates the $i^{th}$ category, and $N$ is the total number of categories in SHADE dataset. $M_i$ is the total number of samples belong to the $i^{th}$ category. $p_i$ and $y_i$ is the prediction and ground-truth of category label respectively. And the indicator function $\mathbbm{1} \{ \cdot, \cdot \}$ is defined as Eq. \ref{indicator}. In the experiments, we gauge the Top-1 Accuracy on every category separately. 
	
	\subsection{Dataset}
	Due to the lack of audio collection recorders on current surveillance cameras, the recent abnormal events detection/classification datasets under real world only contain visual information without ambient sounds. Under this condition, the traditional action classification dataset or abnormal events detection datasets are not suitable for our AVRL framework, so that we choose a synthetic dataset: SHADE \cite{lin2021learning} which is generated in the video game named Grand Theft Auto V (GTA5), including 2149 videos (879,932 frames with the size of $1920\times 1080$). The videos in SHADE are from nine categories, \emph{arrest}, \emph{chase}, \emph{fight}, \emph{knowkdown}, \emph{run}, \emph{shoot}, \emph{scatter}, \emph{normal type 1}, and \emph{normal type 2}. Every category contains about 200 videos with different weather conditions (rain, foggy, sunny, \emph{etc.}) and different happening time. There are 1701 videos (about 80\%) for training, and the remaining 448 videos (about 20\%) for validation or testing. Meanwhile, we extract the audio sounds from videos for the training of audio representation learning module.
	
	\subsection{Experimental Settings}
	In this part, some experimental settings and details are explained. Both the spatial transformations and temporal transformations are imposed on the input video. The input frames are first resized to the size of $240\times 240$, and a random horizontal flip with a probability of 0.5. The temporal transformations sample $50$ frames from the whole frame sequence with a fixed step. The audio signal only preprocess to gain the LMS feature without any other transformations. 
	
	In the training phase, we select Stochastic Gradient Descent (SGD) method as the optimizer. The momentum of SGD optimizer is 0.9, and the weight decay is 0.001. The learning rate is set as $1e-2$ at the beginning, and the learning rate is updated automatically through minimize the validation loss. From the experiment results, the final learning rate is $1e-7$. The batch size in training is set as 12, and we train our model 80 epochs. The experiments run on two NVIDIA GTX 1080TI GPUs.
	
	\begin{table*}[ht]
		\centering
		\small
		\caption{Top-1 accuracy comparison of different approaches on SHADE dataset.}
		\label{results}
		\begin{tabular}{m{0.1\linewidth} m{0.1\linewidth}<{\centering} m{0.1\linewidth}<{\centering} m{0.1\linewidth}<{\centering} m{0.1\linewidth}<{\centering} m{0.1\linewidth}<{\centering} m{0.1\linewidth}<{\centering} c}
			\toprule[1pt]
			Category&MLP& LSTM & LRCN & 3D ResNet & N3D ResNet & AVRL (\textbf{ours}) & $\Delta$ \\
			\hline
			shoot&0.866&0.916&0.905& 0.898 & 0.920&\textbf{0.959} & \textcolor{red}{$\uparrow$}
			\\
			scatter&0.822&0.896&0.896 & 0.905 & 0.923&\textbf{0.954} & \textcolor{red}{$\uparrow$}
			\\
			fight&0.572&0.612&0.630& 0.640 &0.577&\textbf{0.644} & \textcolor{red}{$\uparrow$}
			\\
			arrest&0.643&0.703&0.714& 0.823 &0.837&\textbf{0.900} &\textcolor{red}{$\uparrow$}
			\\
			knokdown&0.663&0.649& 0.621&0.771& 0.742&\textbf{0.808} & \textcolor{red}{$\uparrow$}
			\\
			run&0.627 & 0.650 & 0.648&0.606 &\textbf{0.646}&0.595 & \textcolor{blue}{$\downarrow$}
			\\
			chase&0.582 & 0.619 & 0.641& 0.578&\textbf{0.660}&0.551 &  \textcolor{blue}{$\downarrow$}
			\\
			normal1&0.611 &  0.635 & 0.627&0.656&0.671&\textbf{0.700} & \textcolor{red}{$\uparrow$}
			\\
			normal2& 0.712 & 0.718 & 0.750& 0.787&\textbf{0.804}&\textbf{0.804} & -
			\\
			\toprule[1pt]
		\end{tabular}
	\end{table*}

	\subsection{Performance on SHADE Dataset}
	Table \ref{results} illustrates the comparisons among listed models. Specially, we compare the performance with several visual-based video classification models, such as MLP, LSTM \cite{DBLP:conf/interspeech/SakSB14}, LRCN \cite{DBLP:journals/pami/DonahueHRVGSD17}, 3D ResNet \cite{DBLP:conf/iccvw/HaraKS17}, and N3D ResNet \cite{lin2021learning}. The comparison demonstrates that the audio information improves the classification accuracy significantly. The model introducing 3D convolutional layer, \emph{e.g.} 3D ResNet and N3D ResNet, achieves the better performance among visual-based methods. Compared to the state-of-the-art method, N3D ResNet, on SHADE dataset, the proposed AVRL framework achieve higher Top-1 accuracy on majority of the categories, and only two categories, run and chase, are lower than N3D ResNet. Our AVRL embeds the audio representation learning module into the visual-based method 3D ResNet with an extremely simple fusion module, reaching the notable advance on Top-1 accuracy for events classification. The events with special ambient sounds, such as \emph{shoot}, \emph{scatter}, \emph{arrest}, and \emph{knokdown}, have the remarkable increase, which means the AVRL framework learns the discriminating feature in audio signals and aids the classification effectively. Meanwhile, only a simple concatenation method to fuse the audio and visual representations can gain a great improvement, showing the superiority of the multi-modal learning in abnormal events detection.
	
	\begin{figure*}[h]
		\centering
		\includegraphics[width=\linewidth]{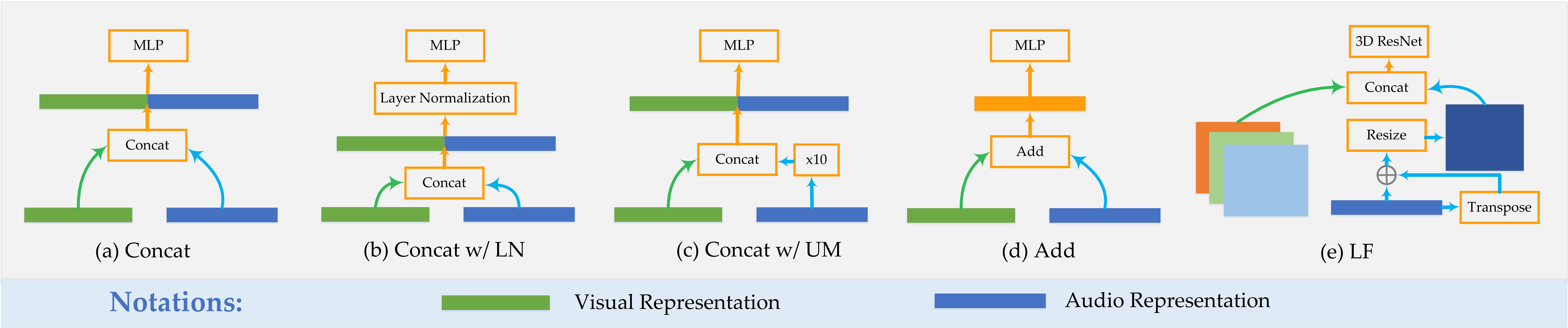}
		\caption{The compared Audio-Visual Representation Fusion modules. Through the experimental comparison, we select the Concat strategy eventually in proposed AVRL.}
		\label{FusionStrategy}
	\end{figure*}

	\subsection{Comparison of Fusion Strategy}
	\label{comp_fusion}
	In Section \ref{method}, we only introduce the concatenation fusion strategy. Actually, we design several different fusion strategy and compare the performance, and we choose the concatenation fusion strategy (Concat) finally. Besides the Concat strategy, the other fusion strategies are concatenation with layer normalization (Concat w/ LN),  concatenation with unified magnitude (Concat w/ UM), addition fusion strategy (Add), and low-level feature fusion strategy (LF).
	
	From the experiments, we find that the audio representation has a lower magnitude than video representation. The average visual representation is tenfold to that of audio representation. We attempt to introduce the layer normalization and a scale factor $S$ respectively to alleviate this divergence, acquiring the Concat w/ LN and Concat w/ UM strategies. Meanwhile, to find out the suitable fusion strategy, we design different prevalent feature fusion strategies, Add and LF, to select a better fusion strategy.

	The details of several fusion strategy mentioned above are shown in Fig. \ref{FusionStrategy}. First, the Concat ((a) in Fig. \ref{FusionStrategy}) directly concatenates the visual and audio representation along channel dimension. The Concat w/ LN introduces the Layer Normalization after feature concatenation to balance the visual representation and audio representation. In addition, we adjust the magnitude by multiplying an hand-craft scale factor $S=10$ to the audio representation in Concat w/ UM. Due to the same shape of visual and audio representations, we naturally add the audio feature to visual feature, fusing the multi-modal features. The former fusion strategies all fuse the representations in the high-level feature. In contrast, we embed the audio representation in the low-level feature from visual representation. We first calculate the inherent correlation in audio representation by outer product, gaining the audio correlation map $Corr_{a} \in \mathbb{R}^{1\times 512\times 512}$. And the $Corr_{a}$ is resized to the same spatial shape with the input frames through bilinear interpolation. Then, the resized $Corr_{a}$ is stacked with all of input video frames along channel dimension. 
	
	\begin{figure*}[ht]
		\centering
		\subfigure{
			\begin{minipage}[b]{0.5\textwidth}
				\centering
				{\scriptsize Concat} \vspace{1pt} \\
				\includegraphics[width=\textwidth]{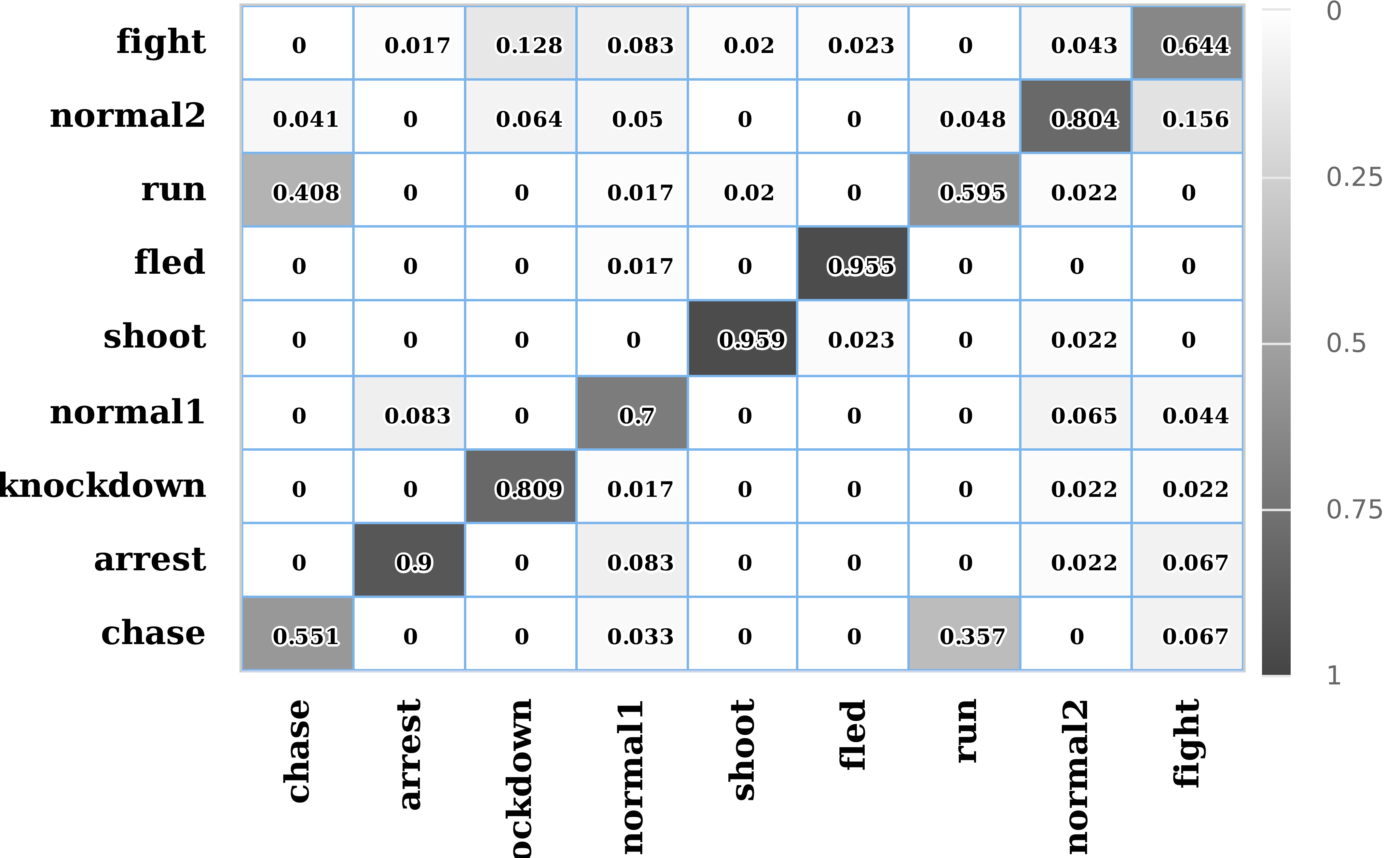}
			\end{minipage}
			\begin{minipage}[b]{0.5\textwidth}
				\centering
				{\scriptsize Concat w/ LN}  \vspace{1pt} \\
				\includegraphics[width=\textwidth]{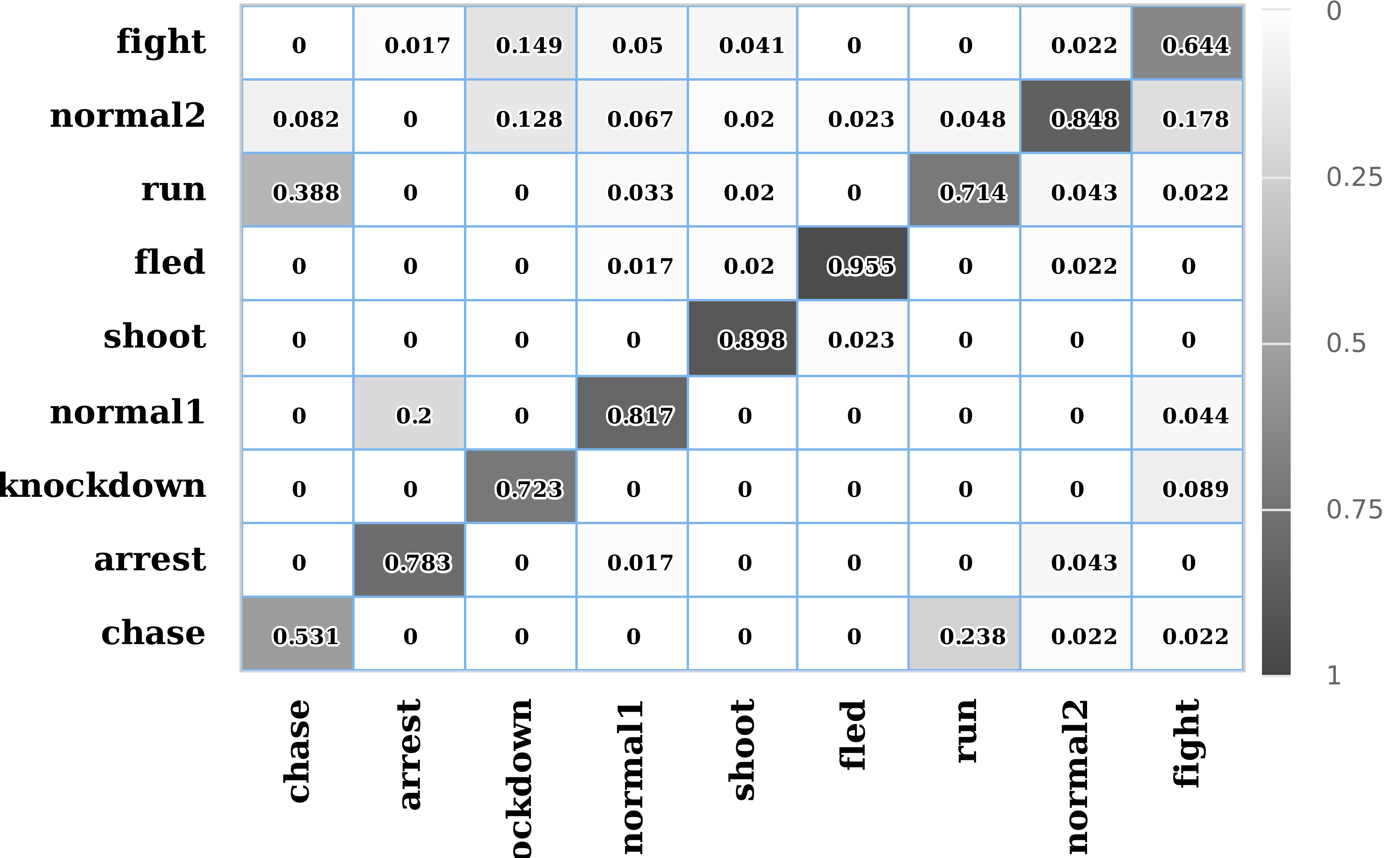}
			\end{minipage}
		}
		\subfigure{
			\begin{minipage}[b]{0.5\textwidth}
				\centering
				{\scriptsize Concat w/ UM} \vspace{1pt} \\
				\includegraphics[width=\textwidth]{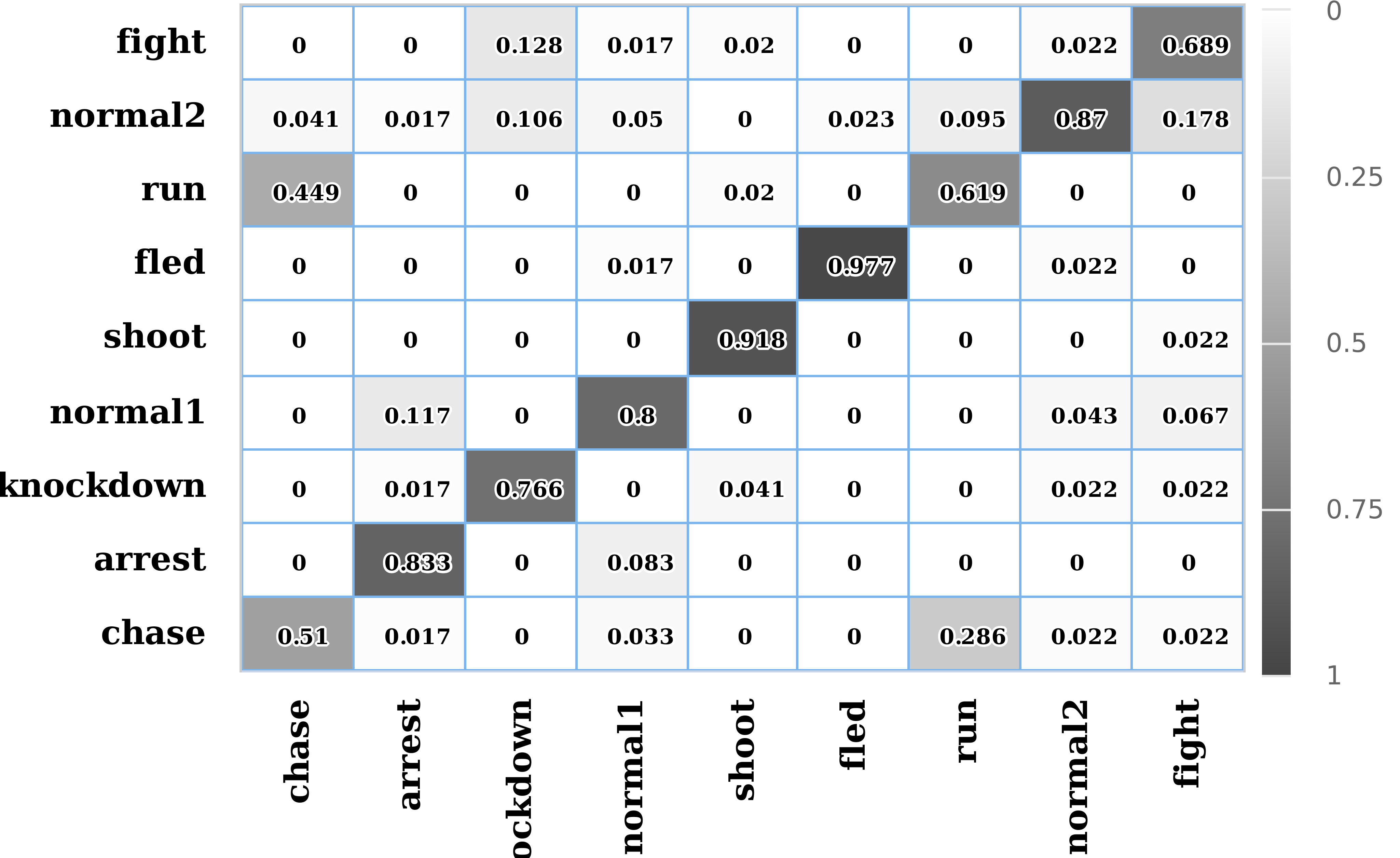}
			\end{minipage}
			\begin{minipage}[b]{0.5\textwidth}
				\centering
				{\scriptsize Add}  \vspace{1pt} \\
				\includegraphics[width=\textwidth]{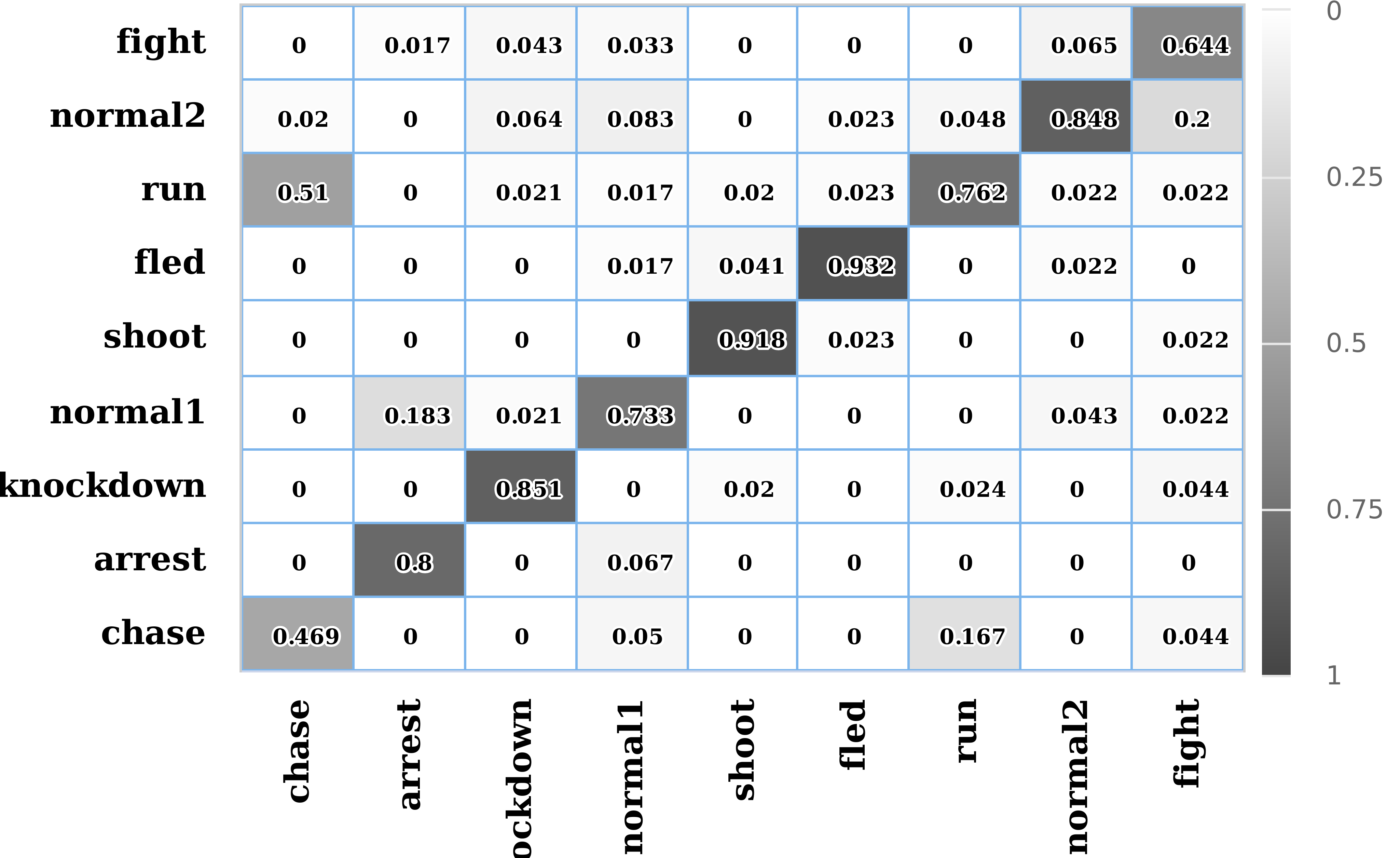}
			\end{minipage}
		}
		\subfigure{
			\begin{minipage}[b]{0.5\textwidth}
				\centering
				{\scriptsize LF} \vspace{1pt} \\
				\includegraphics[width=\textwidth]{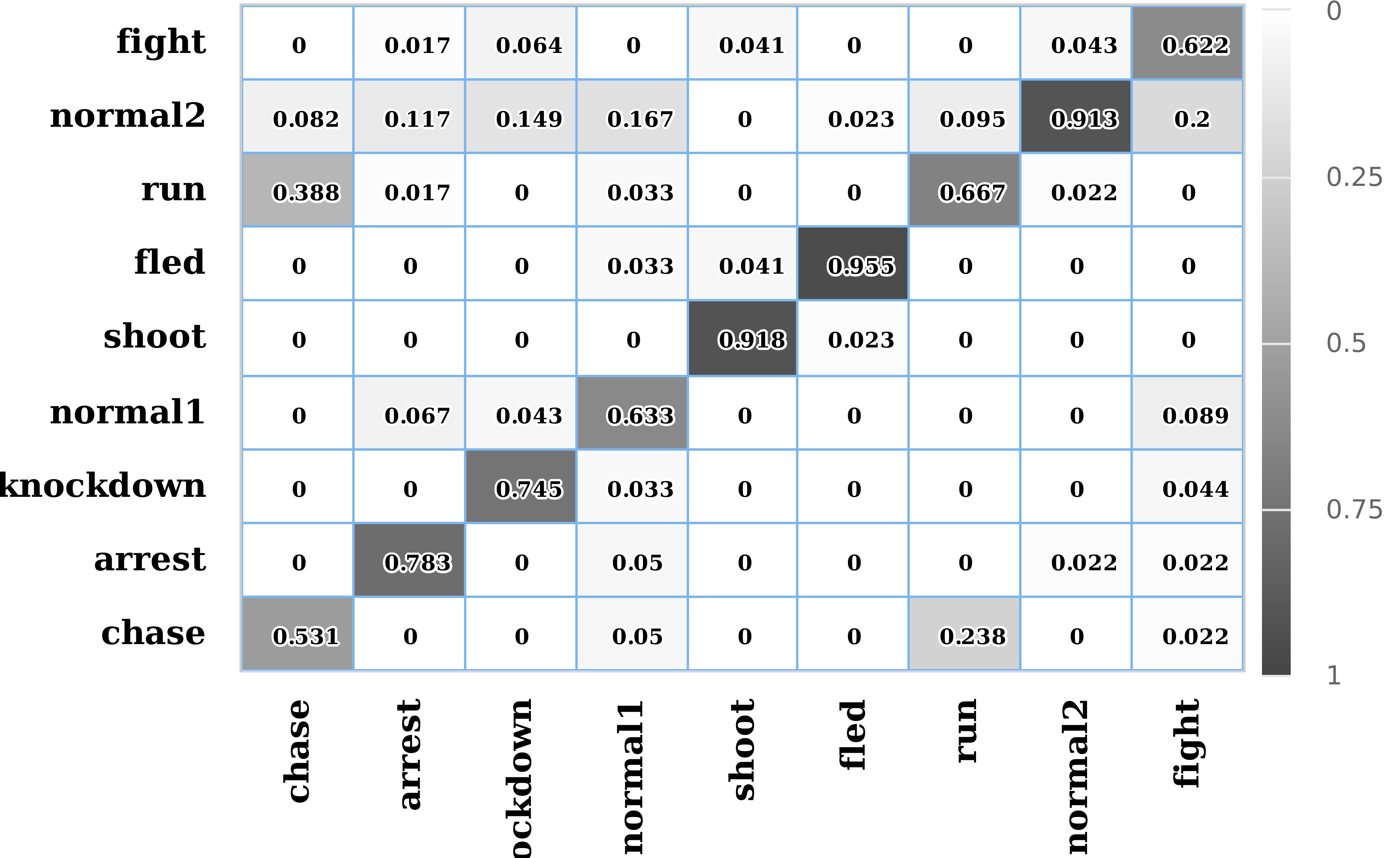}
			\end{minipage}
		}
		\caption{The visualization of classification results from the compared fusion strategies. Rectangles consisting of 9$\times$9 grids are classification results in 9 categories.}
		\label{confuse}
	\end{figure*}
	
	\begin{table*}[h]
		\centering
		\small
		\caption{Top-1 accuracy comparison of different approaches on SHADE dataset.}
		\label{fusion}
		\begin{tabular}{m{0.1\linewidth} m{0.1\linewidth}<{\centering} m{0.15\linewidth}<{\centering} m{0.15\linewidth}<{\centering} m{0.1\linewidth}<{\centering} m{0.1\linewidth}<{\centering}}
			\toprule[1pt]
			Category&Concat& Concat w/ LN & Concat w/ UM & Add & LF  \\
			\hline
			shoot&\textbf{0.959}&0.898&0.918& 0.918 & 0.918
			\\
			scatter&0.954&0.954&\textbf{0.977} & 0.931 & 0.954
			\\
			fight&0.644&0.644&\textbf{0.689}& \textbf{0.689} &0.644
			\\
			arrest&\textbf{0.900}&0.783&0.833& 0.800 &0.783
			\\
			knokdown&0.808&0.723& 0.766&\textbf{0.851}& 0.745
			\\
			run&0.595 & 0.714 & 0.619&\textbf{0.762} &0.666
			\\
			chase&\textbf{0.551} & 0.531 & 0.510& 0.469&0.531
			\\
			normal1&0.700 &  \textbf{0.816} & 0.800&0.733&0.633
			\\
			normal2& 0.804 & 0.848 & 0.869& 0.848&\textbf{0.913}
			\\
			\toprule[1pt]
		\end{tabular}
	\end{table*}
	
	The comparison of fusion strategy introduced above is illustrated in Table \ref{fusion}, and the classification results are demonstrated in Fig. \ref{confuse}. From the comparison, the following conclusions are obtained: (1) The Concat strategy achieves relatively high performance among several fusion strategies, reaching the top place on  \emph{shoot}, \emph{arrest}, and \emph{chase}. (2) Compared to Concat, the Concat w/ LN reaches higher accuracy on \emph{run}, \emph{normal1}, and \emph{normal2}, but the performances on events with discriminating ambient sounds (\emph{shoot} and \emph{arrest}) have significant decrease, which means the Concat w/ LN can not exploit the ambient sounds sufficiently. (3) The Concat w/ UM achieves better performance than Concat w/ LN, and the accuracy on \emph{run} has increase, but the accuracy on \emph{chase} decrease notably. We find that the Concat w/ UM strategy tends to consider the events with running people as \emph{run} events, causing the increase of accuracy on \emph{run}. However, the \emph{chase} events contain two running people in scenes, and the  Concat w/ UM confuses \emph{chase} with \emph{run}, leading to the decrease of accuracy on \emph{chase}. (4) The Add also achieve high performances on several events, but the improvements is not higher than Concat, and there is a significant decrease of performance on \emph{chase} due to the same reason as (3). Finally, the improvement of LF strategy is not obvious than other fusion strategies. In summary, with the comprehensive consideration, we select the Concat strategy in audio-video representation fusion module.
	
	\section{Discussion}
	\label{dis}
	To further verify the advantage of AVRL and elaborate the principle about multi-modal learning in AVRL, some discussions are developed in this section. 
	
	\subsection{The performance on Dark Scenes}
	In the real world, the abnormal events usually happen in the dark environment. One of the motivations of AVRL is to aid the abnormal events detection task, especially in some degeneration scenes, such as low illumination, noise, \emph{etc.} In order to stand out the robustness of AVRL, we select a subset in SHADE that only includes the dark scenes. The selected dark subset contains 147 events from 18:00 pm to 23:59 pm, 85 events from 0:00 am to 5:59 am, and 145 events occur in daytime with low illumination environments. Furthermore, the comparison between the proposed AVRL and the state-of-the-art N3D ResNet on SHADE dataset. Table \ref{dark} illustrates the comparison on dark subset. From the table, we find that the proposed AVRL framework surpasses the state-of-the-art method N3D ResNet absolutely. 
	
	\begin{table}[h]
		\centering
		\small
		\caption{Top-1 accuracy comparison on dark subset.}
		\label{dark}
		\begin{tabular}{lcc}
			\toprule[1pt]
			Category&N3D ResNet& AVRL (\textbf{ours})\\
			\hline
			shoot&0.832& \textbf{0.977}
			\\
			scatter&0.805& \textbf{1.0}
			\\
			fight&0.551& \textbf{0.961}
			\\
			arrest&0.635& \textbf{1.0}
			\\
			knokdown&0.495&\textbf{1.0}
			\\
			run&0.632 & \textbf{0.925} 
			\\
			chase& 0.591 & \textbf{0.914}
			\\
			normal1&0.628 &  \textbf{0.925} 
			\\
			normal2& 0.754 & \textbf{0.921} 
			\\
			\toprule[1pt]
		\end{tabular}
	\end{table}
	
	The nighttime events in the dark subset usually occur in quite environments, so that the event sounds are discriminative, leading to a better distinguish of anomalies. Due to this point, the classification on \emph{run} and \emph{chase} achieves better accuracy. And some events with notable ambient sounds, such as \emph{shoot}, \emph{scatter}, \emph{arrest}, and \emph{knokdown}, have the accuracy extremely close to 1.0. Interestingly, the AVRL does not reaches high performance on \emph{fight}, \emph{run}, and \emph{chase} in SHADE dataset, but the performance on these categories in dark subset increases significantly. The daytime samples have noise from crowds involving chatting, bustle, \emph{etc.} which do harm to the assistance from audio. Overall, the AVRL framework addresses the anomaly detection with dark scenes effectively, proving the advantage of audio-visual multi-modal learning. 
	
	\subsection{Ablation Study}
	The ablation study is conduct to further compare the influences from audio representation and visual representation. We consider three schemes: only visual scheme (3D ResNet), only audio (VGGish), and audio-visual (the proposed AVRL). The results of ablation study are shown in Table \ref{ablation}.
	
	\begin{table}[h]
		\centering
		\small
		\caption{The comparison on top-1 accuracy in ablation study.}
		\label{ablation}
		\begin{tabular}{lccc}
			\toprule[1pt]
			Category&3D ResNet& VGGish & AVRL (\textbf{ours})\\
			\hline
			shoot&0.898&0.918 & \textbf{0.959}
			\\
			scatter&0.905 & 0.591& \textbf{0.954}
			\\
			fight&0.640 &0.133 & \textbf{0.644}
			\\
			arrest& 0.823& 0.417 & \textbf{0.900}
			\\
			knokdown&0.771 & 0.638 & \textbf{0.808}
			\\
			run& \textbf{0.606} & 0.476 & 0.595
			\\
			chase& \textbf{0.578} & 0.326 & 0.551
			\\
			normal1&0.656& 0.433 &  \textbf{0.700}
			\\
			normal2& 0.787 & 0.326 & \textbf{0.804}
			\\
			\toprule[1pt]
		\end{tabular}
	\end{table}
	
	From the Table \ref{ablation}, the following conclusion can be obtained: (1) The 3D ResNet (only visual) achieves better accuracy than VGGish (only audio), which means the visual representation makes heavier contributions on abnormal events detection than audio representation. (2) VGGish (only audio) reaches high classification accuracy on the events with special ambient sounds, especially the \emph{shoot}. (3) Although the performance of VGGish (only audio) is weak, the AVRL (audio-visual) achieves better classification accuracy by combining the audio and video representations. 
	
	\begin{figure}[!h]
		\centering
		\subfigure[Temporal Patch]{
			\label{temporal_patch}
			\includegraphics[width=\linewidth]{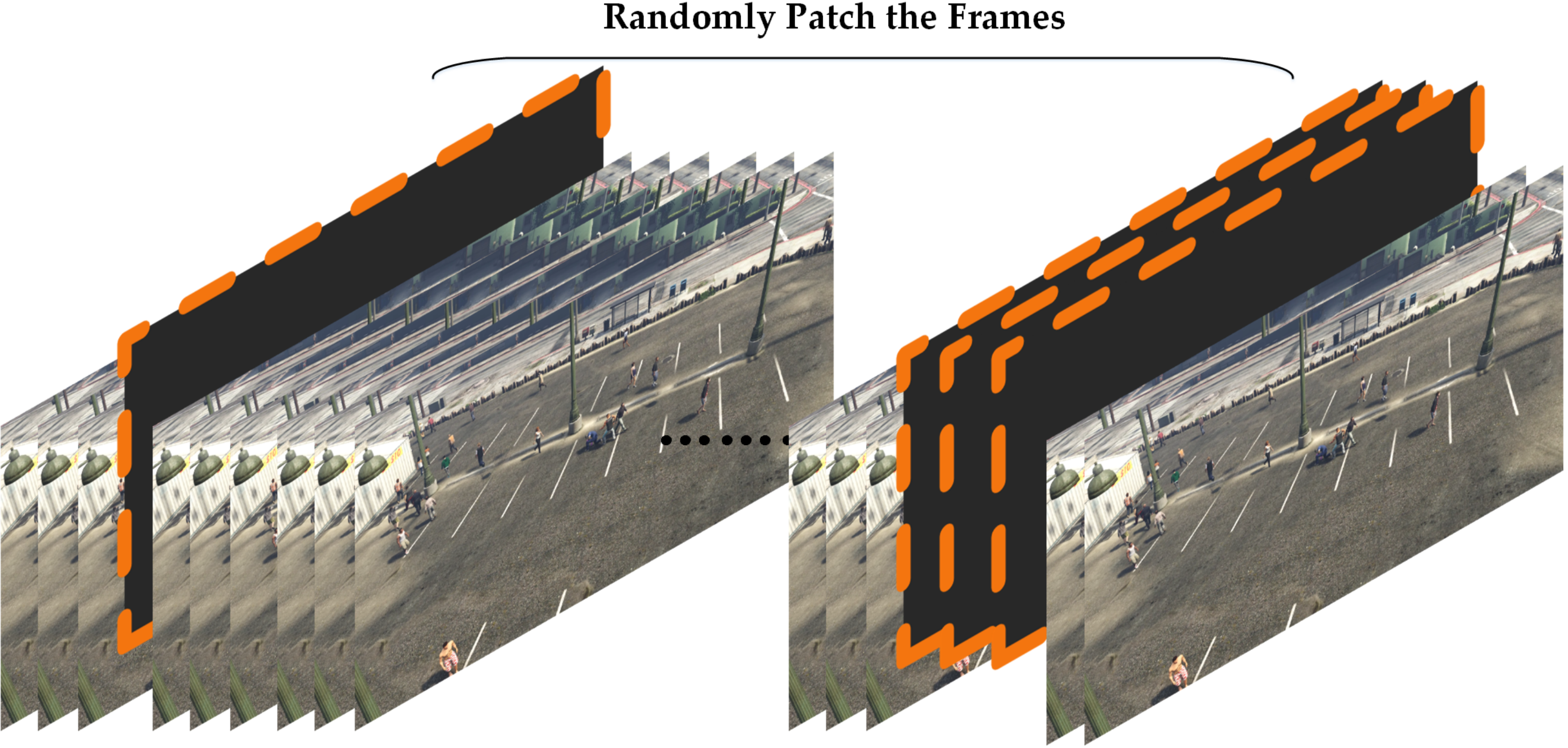}
		}
		\subfigure[Spatial Patch]{
			\label{spatial_patch}
			\includegraphics[width=\linewidth]{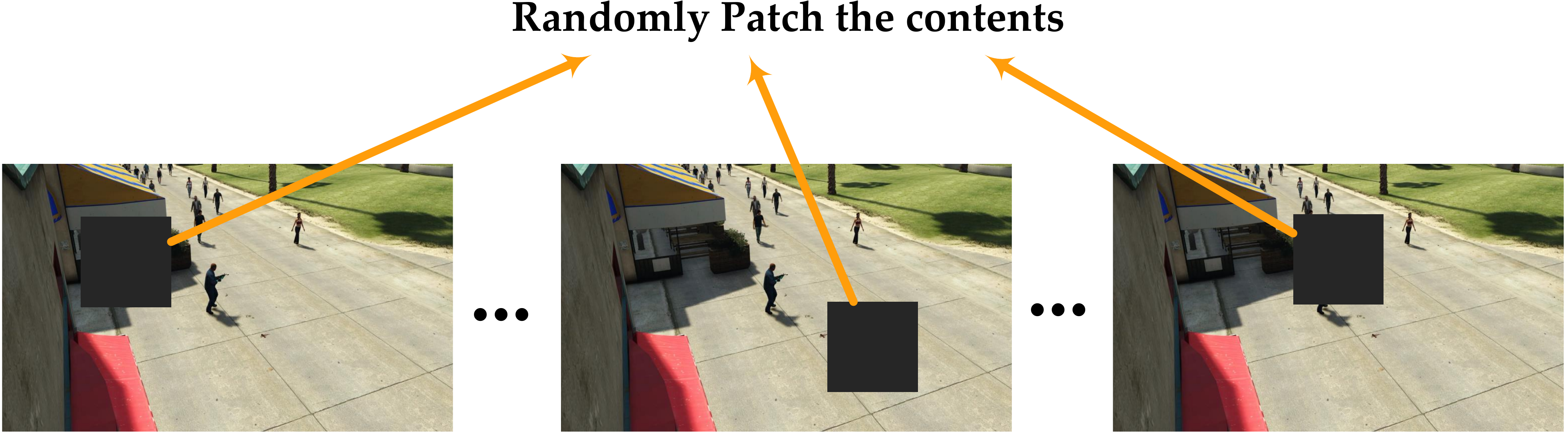}
		}
		\caption{The temporal and spatial patch applied in this section.}
		\label{show_patch}
	\end{figure}
	
	\begin{figure}[!h]
		\centering
		\subfigure[Temporal Patch]{
			\label{temporal}
			\includegraphics[width=\linewidth]{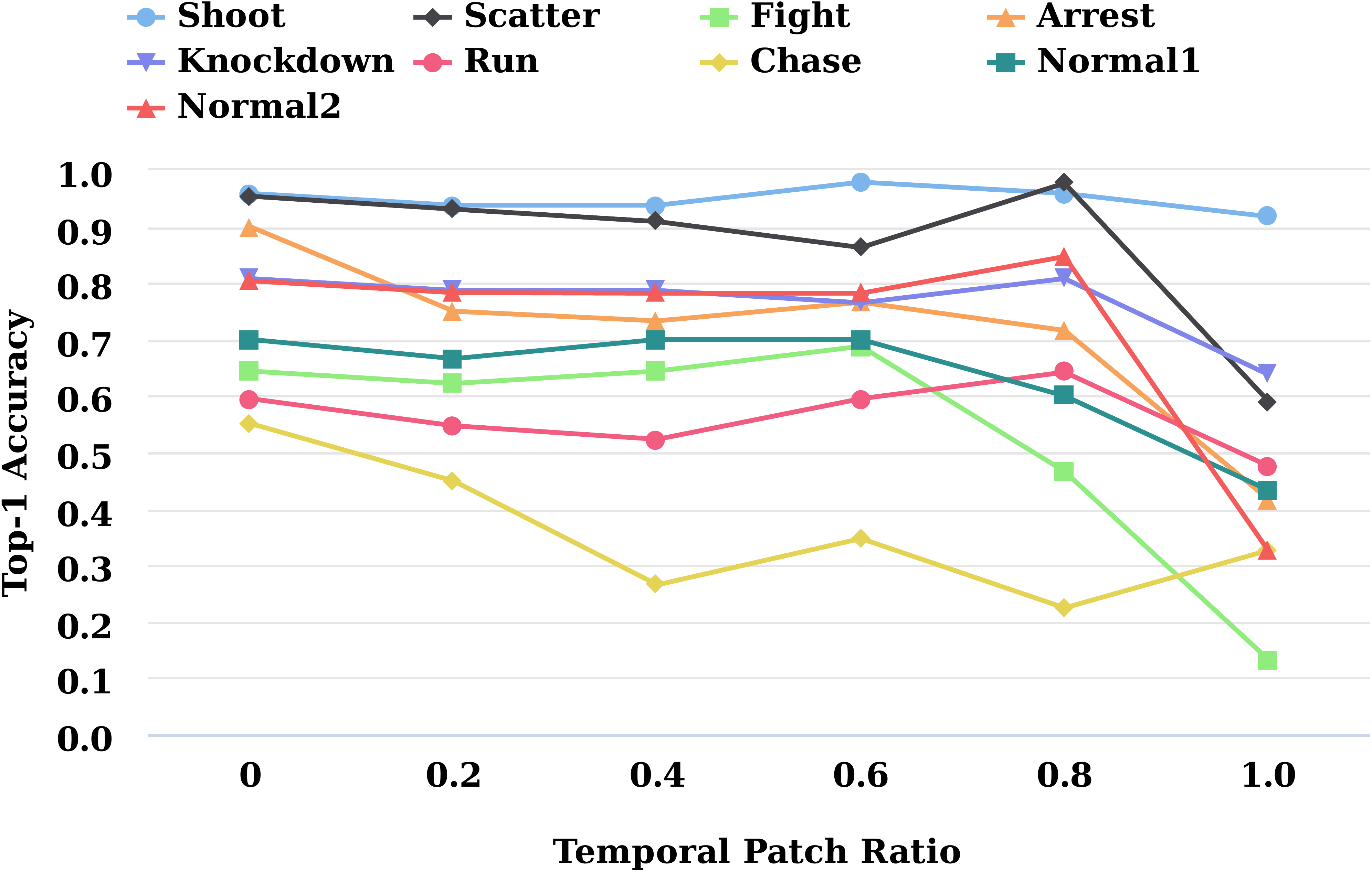}
		}
		\subfigure[Spatial Patch]{
			\label{spatial}
			\includegraphics[width=\linewidth]{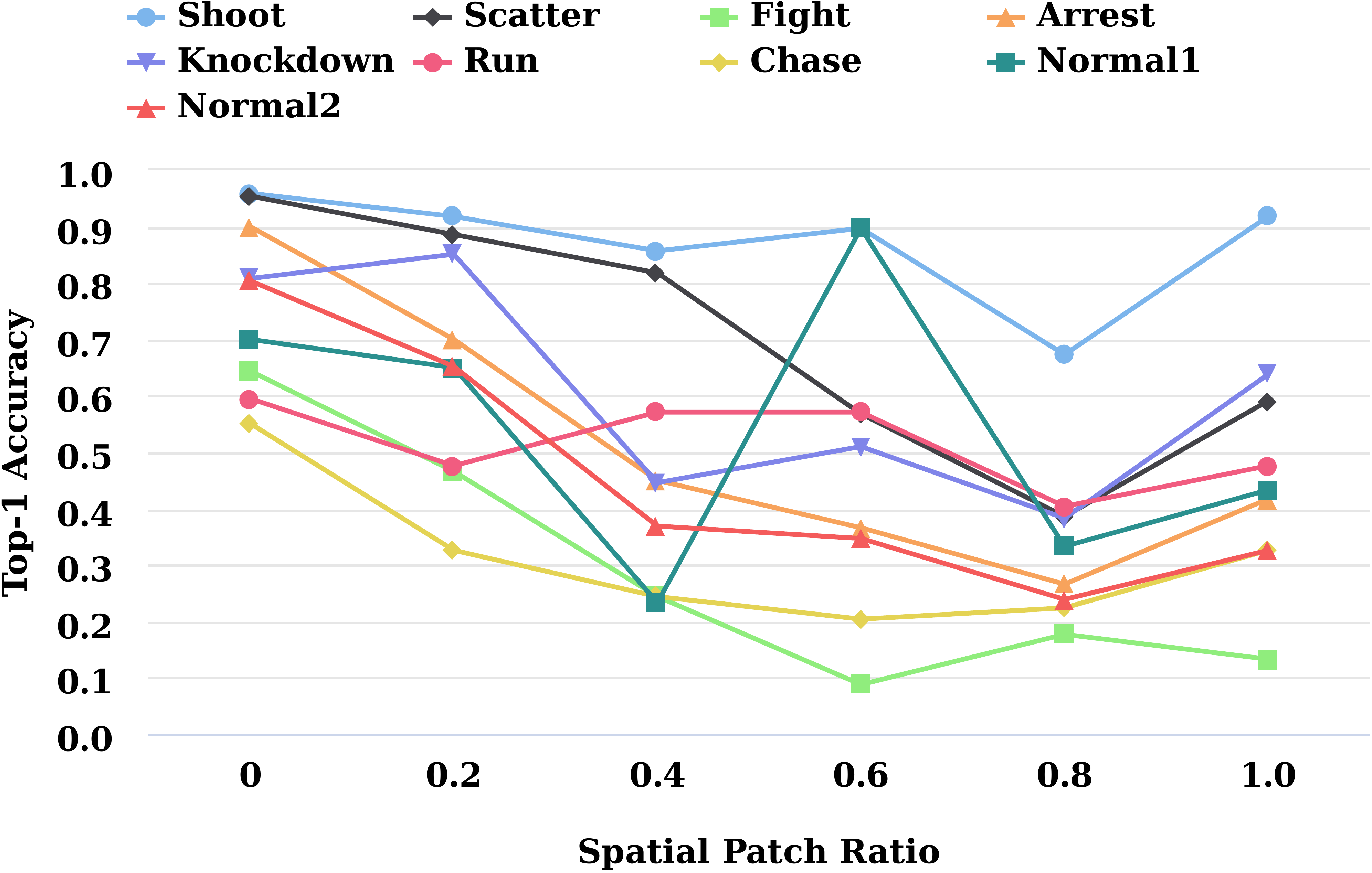}
		}
		\caption{The influence on classification accuracy of temporal and spatial patch ratio.}
		\label{patch}
	\end{figure}
	
	\subsection{Temporal and Spatial Patches}
	In order to probe the relationship between visual representation and audio representation, we add patches randomly in videos, as shown in Fig. \ref{show_patch}. As for temporal patch, the video frames are randomly selected under a proportion and colored by black (Fig. \ref{temporal_patch}). Moreover, the black patches with a proportion of the full spatial area are put to random positions on every frames (Fig. \ref{spatial_patch}). The comparison results are demonstrated in Fig. \ref{temporal} and Fig. \ref{spatial}.

	With the increase of patch ratio, the classification accuracy of both temporal patch and spatial patch decreases unsurprisingly. The effect of temporal patch is less than that of spatial patch, so that the accuracy of spatial patch have fallen more against temporal patch, which means that the spatial correlation in single frame provide more contributions than temporal correlation among frames for events classification. From Fig. \ref{patch}, we find that the accuracy on some categories with 80\% spatial patch ratio is lower than that of the method with only audio representation. This probably because the high spatial patch ratio lead to the extremely noisy distortion on content and this distortion mislead the visual representation learning, causing the impediment to events detection. 
	
	\section{Conclusion}
	\label{concl}
	This paper proposes a novel multi-modal learning framework AVRL for abnormal events detection. The AVRL framework attempts to improve the abnormal events detection accuracy in crowd scenes through combining the visual representation and audio representation from the surveillance videos. Specifically, a 3D ResNet is applied for visual representation learning and a audio analysis model, VGGish, is used for audio representation learning. On obtaining the audio and visual representations, a simple concatenation feature fusion module is exploited to fuse the multi-modal representations. Experiments show that the proposed AVRL has significant superiority compared with state-of-the-art visual-based methods. 
	
	In the AVRL, we attempt several feature fusion strategies preliminarily and the fusion module still needs optimize to achieve better anomaly detection accuracy. Meanwhile, due to the limitation of datasets, the experiments are only conducted on a synthetic dataset, SHADE dataset, rather than real-world dataset. With the emergence of surveillance video datasets with audio signals, we will attempt to pursue this research on more datasets.

	
	%

	%
	%
	%
	%
	%

	\ifCLASSOPTIONcaptionsoff
	\newpage
	\fi

	
	
	\bibliographystyle{IEEEtran}
	\bibliography{IEEEabrv,strings}
	%
	%
	%
	
	%

	
	

\end{document}